\documentclass[10pt,twocolumn,letterpaper]{article}

\usepackage[pagenumbers]{cvpr} 
\usepackage{graphicx}
\usepackage{amsmath}
\usepackage{amssymb}
\usepackage{booktabs}
\usepackage{textcomp}
\usepackage[T1]{fontenc}

\usepackage{hyperref}
\hypersetup{hidelinks,
	colorlinks=true,
    urlcolor = blue,
	pdfstartview=Fit,
	breaklinks=true}

\usepackage[dvipsnames]{xcolor}
\usepackage[capitalize]{cleveref}
\crefname{section}{Sec.}{Secs.}
\Crefname{section}{Section}{Sections}
\Crefname{table}{Table}{Tables}
\crefname{table}{Tab.}{Tabs.}

\def\cvprPaperID{477} 
\def\confName{CVPR}
\def\confYear{2023}

\begin{document}

\title{Neuron Structure Modeling for Generalizable Remote Physiological Measurement}

\author{Hao LU$^{1,2}$, Zitong YU$^{3}$, Xuesong NIU$^{4}$, Yingcong CHEN$^{1,2,}$\thanks{Corresponding author.}\\
$^{1}$The Hong Kong University of Science \& Technology (Guangzhou), \\
$^{2}$The Hong Kong University of Science \& Technology,
$^{3}$Great Bay University, $^{4}$Kuaishou Technology\\
hlu585@connect.hkust-gz.edu.cn, zitong.yu@ieee.org,\\
nxsedson@gmail.com, yingcongchen@ust.hk
}

\maketitle


\begin{abstract}
Remote photoplethysmography (rPPG) technology has drawn increasing attention in recent years. It can extract Blood Volume Pulse (BVP) from facial videos, making many applications like health monitoring and emotional analysis more accessible. However, as the BVP signal is easily affected by environmental changes, existing methods struggle to generalize well for unseen domains. In this paper, we systematically address the domain shift problem in the rPPG measurement task. We show that most domain generalization methods do not work well in this problem, as domain labels are ambiguous in complicated environmental changes. In light of this, we propose a domain-label-free approach called NEuron STructure modeling (NEST). NEST improves the generalization capacity by maximizing the coverage of feature space during training, which reduces the chance for under-optimized feature activation during inference. Besides, NEST can also enrich and enhance domain invariant features across multi-domain. We create and benchmark a large-scale domain generalization protocol for the rPPG measurement task. Extensive experiments show that our approach outperforms the state-of-the-art methods on both cross-dataset and intra-dataset settings. The codes are available at \href{https://github.com/LuPaoPao/NEST}{https://github.com/LuPaoPao/NEST}. 
\end{abstract}

\vspace{-5mm}
\section{Introduction}
\label{sec:intro}
Physiological signals such as heart rate (HR), and heart rate variability (HRV), respiration frequency (RF) are important body indicators that serve not only as vital signs but also track the level of sympathetic activation~\cite{sun2016lstm,kessler2017pain,mcduff2022applications}. Traditional physiological measurements, such as electrocardiograms, heart rate bands, and finger clip devices, have high accuracy. However, they are costly, intrusive, and uncomfortable to wear for a long time. Remote photoplethysmography (rPPG) can extract blood volume pulse (BVP) from face video, which analyzes the periodic changes of the light absorption of the skin caused by heartbeats. Then various physiological indicators (such as HR and HRV) can be calculated based on BVP signals~\cite{yu2019remote,CVD2020,yu2021physformer}. With non-intrusion and convenience, the rPPG-based physiological measurement method can only use an ordinary camera to monitor physiological indicators and gradually become a research hotspot in the computer vision field~\cite{mcduff2021camera,2020PulseGAN,lu2021dual,bvpnet,niu2019robust,Qiu2019EVM}.

\begin{figure}[!t]
\begin{center}
\includegraphics[width=1.0\linewidth]{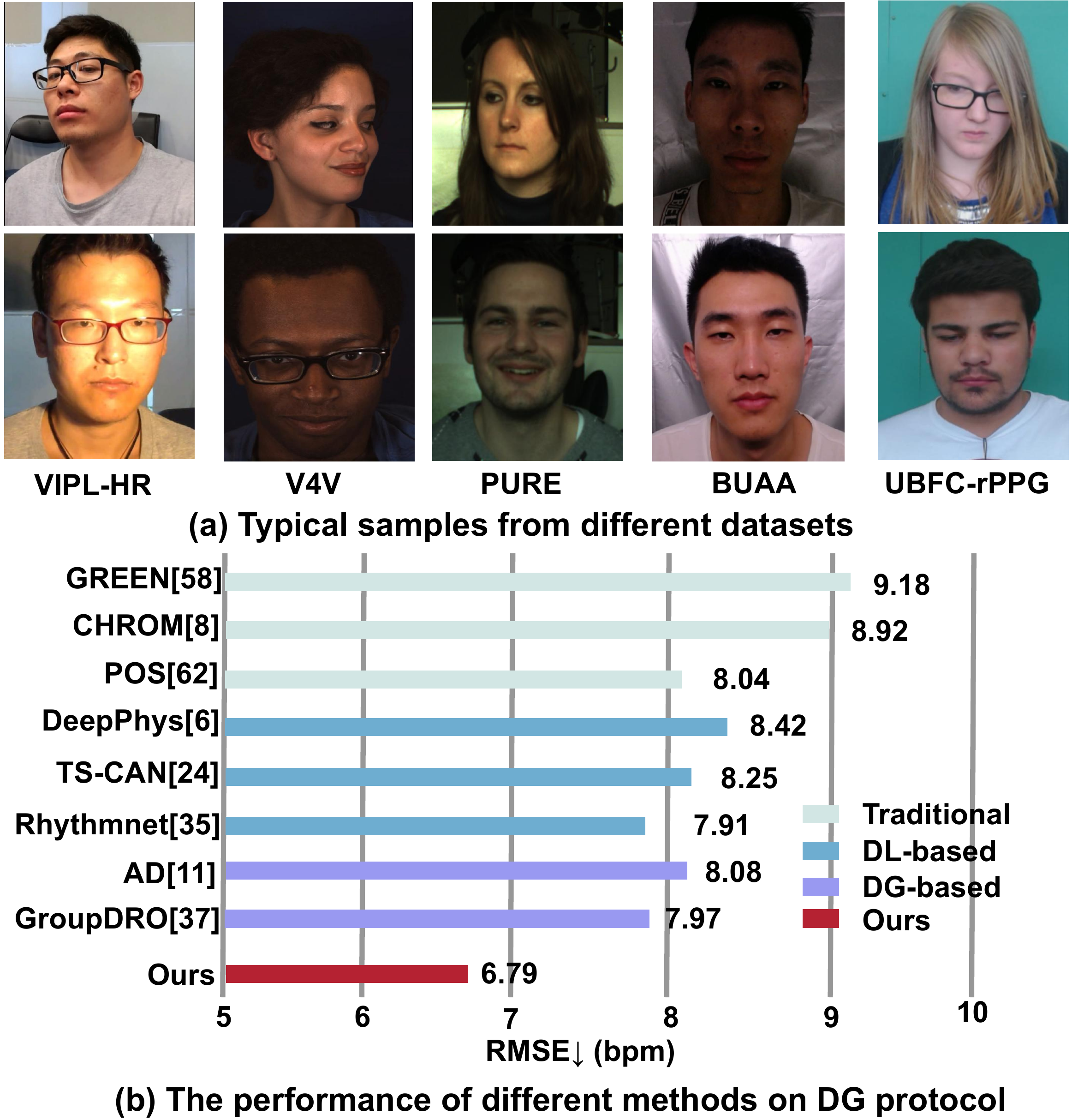}
\vspace{-8mm}
\caption{(a) Typical samples from different publicly rPPG datasets: VIPL-HR~\cite{niu2019robust}, V4V~\cite{revanur2021V4V}, UBFC-rPPG~\cite{UBFC2017}, BUAA~\cite{xi2020BUAA}, PURE~\cite{PURE2014}. (b) The performance of different methods on DG protocol (test on the UBFC-rPPG dataset with training on the VIPL, V4V, PURE, and BUAA).}
\label{fig:Intro}
\vspace{-11.5mm}
\end{center}
\end{figure}

Traditional rPPG measurement methods include signal blind decomposition~\cite{poh2010non,2014ICA,2011rPPGPCA} and color space transformation~\cite{verkruysse2008remote_GREEN,de2013robust_CHROM,wang2017algorithmic_POS}. These approaches rely on heartbeat-related statistical information, only applicable in constrained environments. In recent years, deep learning (DL) based  approaches~\cite{Chen_2018_ECCV,liu2020multi,yu2019remote,yu2021physformer,lee2020meta,niu2019robust,song2020heart,Qiu2019EVM,lu2021dual} have shown their great potentials in rPPG measurement. By learning dedicated rPPG feature representation, these methods achieve promising performance in much more complicated environments~\cite{niu2019robust,revanur2021V4V,PURE2014,UBFC2017,xi2020BUAA}.

However, deep learning methods suffer from significant performance degradation when applied in real-world scenarios. This is because most training data are captured in lab environments with limited environmental variations. With domain shifts (e.g., different illumination, camera parameters, motions, etc.), these models may struggle to generalize for the unseen testing domain. To validate this, we conduct a cross-dataset evaluation shown in Fig. \ref{fig:Intro}(b). As shown, all DL-based methods do not work well in this evaluation. Furthermore, it is worth noting that DeepPhys~\cite{Chen_2018_ECCV} and TS-CAN~\cite{liu2020multi} even perform inferior to traditional approach POS~\cite{wang2017algorithmic_POS}. 

\begin{figure*}[!ht]
\begin{center}
\includegraphics[scale=0.5]{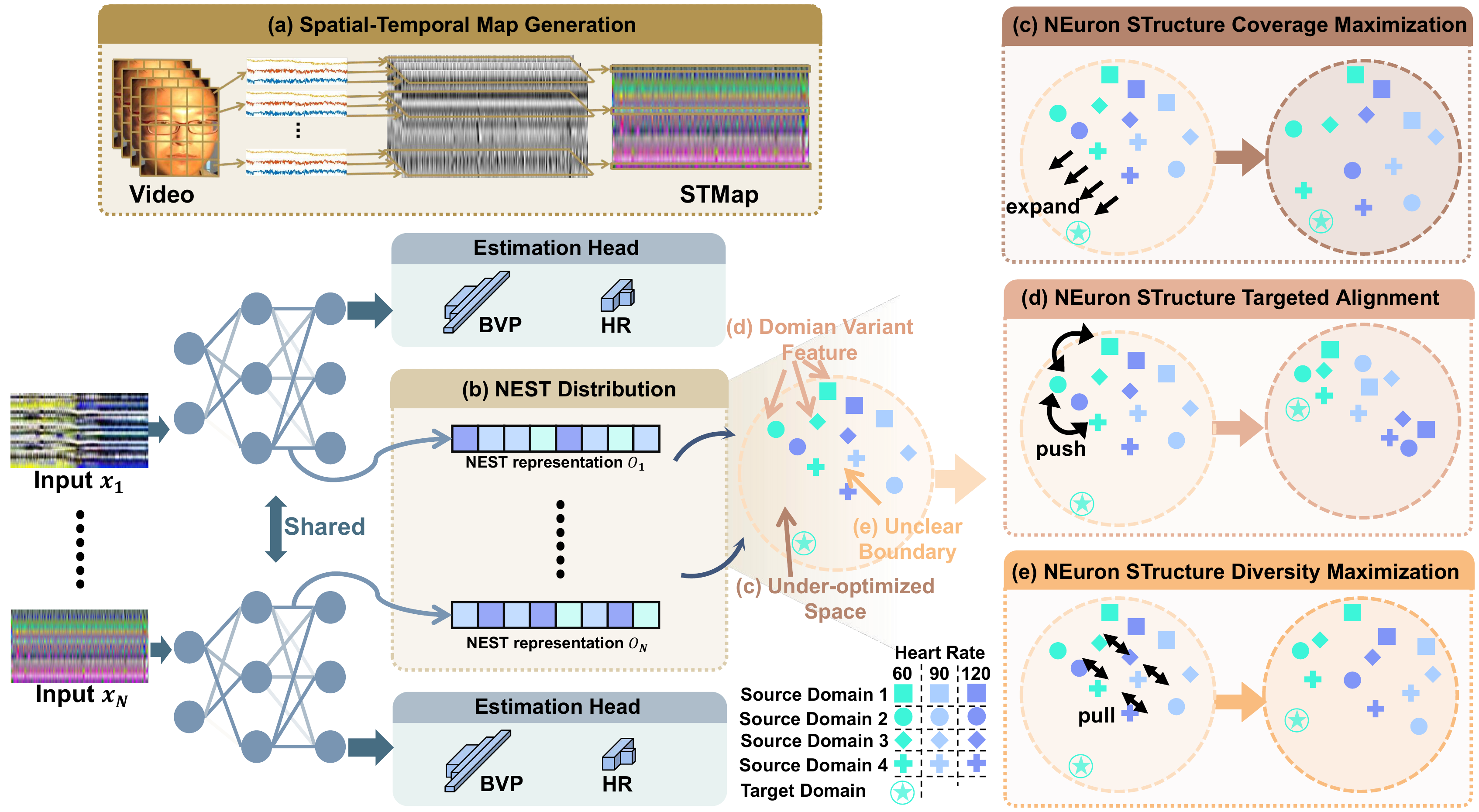}
\end{center}
\vspace{-5mm}
\caption{An overview of the proposed method. (a) Spatial-temporal map (STMap) is extracted from face video~\cite{niu2019robust}. (b) NEST representation is calculated from the samples of the different domains. (c) The under-optimized space in NEST presentation distribution is narrowed by NEST-CM. (d) NEST-TA align the feature of multi domain at instance-level. (e) The distance of different samples are pulled away by NEST-DM.}
\label{fig:Framework}
\vspace{-5mm}
\end{figure*}

To improve the performance on the unseen domains, one common practice is to incorporate Domain Generalization (DG) approaches, e.g., encouraging intermediate features to be domain-invariant~\cite{wang2022DG,ganin2015AD,VREx,NC2022PAMI,GroupDRO,lv2022causality}. However, as shown in Fig.~\ref{fig:Intro} (b) (AD~\cite{ganin2015AD} and GroupDRO~\cite{GroupDRO}), the improvements are still quite limited. One reason is that existing DG methods assume that domain labels can be clearly defined (e.g., by data source). Unfortunately, different background scenes, acquisition devices, or even identities could cause very different data distributions in rPPG measurement tasks, and such distribution discrepancy may exist within or cross datasets. In this case, explicitly defining domains is very difficult, and simply treating one dataset as one domain may lead to inferior performance. This is termed as \textit{agnostic domain generalization}. 
Besides this, as physiological cues are usually much more subtle than the various noise, the model may overfit in the source domain with the limited training data.

In this paper, we propose the NEural STructure modeling (NEST), a principled solution to the abovementioned problems. The main idea of NEST is to narrow the under-optimized and redundant feature space, align domain invariant features, and enrich discriminative features. Our intuition is as follows. Neural structure refers to the channel activation degree in each convolution layer, which reveals the discriminative feature combination for the specific sample. 
As the limited variation in a certain restricted domain, there are some spaces that the model seldomly is optimized. Out-of-distribution (OOD) samples may cause abnormal activation in these spaces, which may lead to performance degeneration. Therefore, we regularize the neural structure to encourage the model to be more well-conditioned to avoid abnormal activation caused by OOD samples. Specifically, we propose the NEural STructure Coverage Maximization (NEST-CM) that encourages all neural spaces to be optimized during training, reducing the chance of abnormal activation during testing. Secondly, we propose the NEural STructure Targeted Alignment (NEST-TA) that encourage network suppresses domain variant feature by comparing the samples with the similar physiological information. Thirdly, we propose the NEural STructure Diversity Maximization (NEST-DM) to enrich discriminative features against unseen noise. It should be noted that our approach does not rely on domain labels, which is more applicable in the rPPG measurement task. To summarize, our contributions are listed as follows:

1. We are the first to study the domain shift problem in rPPG measurement, which introduces a new challenge, agnostic domain generalization. 

2. We propose the NEural STructure modeling to alleviate domain shift, which is performed by narrowing the under-optimized feature space, and enhancing and enriching domain invariant features. 

3. We establish a large-scale domain generalization (DG) benchmark for rPPG measurement, which is the first DG protocol in this task. Extensive experiments in this dataset show the superiority of our approach.

\vspace{-3mm}
\section{Related work}
\vspace{-1mm}
\subsection{Remote Physiological Measurement}
\vspace{-1mm}
Remote physiological measurement aims to estimate HR and HRV values by analyzing the chrominance changes in the skin. Traditional methods mainly perform color space transformations or signal decomposition to obtain the BVP signal with high SNR~\cite{de2013robust_CHROM,wang2015exploiting,de2014improved,poh2010non}. These methods are designed manually for certain lighting scenes with slight motion and can not generalize to the complex environment in the wild. With powerful modeling capacity, the deep learning (DL) model has been successfully used in remote physiological estimation~\cite{Chen_2018_ECCV,hsu2017deep,synrhythm,niu2019robust,vspetlik2018visual,wang2019vision,yu2019remote}. Among these methods, various model structures are tailored to extract BVP signals from face videos by aggregating spatial and temporal information~\cite{Chen_2018_ECCV,vspetlik2018visual,yu2019remote,CVD2020,yu2021physformer}. Besides, many hand-designed feature maps are used to reduce the computational burden of the network by removing the irrelevant background ~\cite{niu2019robust,song2020heart,hsu2017deep,reiss2019deep,lu2021dual}. Some work utilizes the distribution of the target domain to improve the model performance~\cite{lee2020meta,liu2021metaphys}, which is not practical without touching the unseen domain in advance. Differently, we study the model generalization on domain generalization protocol for rPPG measurement.





\vspace{-1mm}
\subsection{Domain generalization}
\vspace{-1mm}

Domain generalization (DG), i.e., out-of-distribution generalization, aims to generalize to the unseen domain by training a model on multiple or single source domains~\cite{wang2022DG}. The main solution can be categorized into data manipulation, representation learning, and meta-learning. Data manipulation is an important technique to improve the generalization of the model, which can remove irrelevant style information by data generation and data augmentation~\cite{yue2019domain,shankar2018generalizing}. The main goal of representation learning is to learn domain-invariant features by alleviating covariate shift across domain~\cite{ganin2015AD,VREx,NC2022PAMI,GroupDRO}. By simulating domain shift on the virtual dataset, meta-learning seeks transferable and shareable knowledge from multiple domain~\cite{li2018learning,lv2022causality}. Unlike common tasks such as classification, segmentation, and detection~\cite{jia2020single,lehner20223d,li2021semantic,peng2022semantic,lin2021domain}, rPPG measurement tasks do not have explicit boundaries between different domains, i.e., agnostic domain generalization. Recently, some single source domain generalization (SSDG) methods focus on the worst-case formulation without using the domain label~\cite{NC2022PAMI,qiao2020learning,wang2021learning}. Different from all the methods above, we tackle the DG problem of regression task in view of neuron structure modeling.



\section{Method}


\subsection{Overall Framework}

In this work, we propose the NEuron STructure modeling (NSET), without using domain label, to alleviate the performance degradation caused by domain shift as illustrated in Fig.~\ref{fig:Framework}. Specifically, we first compressed the face video into the spatial-temporal map (STMap) as the input of CNN refer to ~\cite{niu2019robust} as shown in Fig.~\ref{fig:Framework} (a). Sequentially, we define an Activated NEuron STructure (NEST) representation that describes the activated degree of different channels of the network as shown in Fig.~\ref{fig:Framework} (b). To alleviate the domain shift, we model a unified NEST distribution by three regularizations: (1) NEuron STructure Coverage Maximization (NEST-CM) is used to narrow under-optimized space that is redundant and unstable as shown in Fig.~\ref{fig:Framework} (c); (2) NEuron STructure Targeted Alignment (NEST-TA) is introduced to align multi-domain feature distribution at instance-level as shown in Fig.~\ref{fig:Framework} (d). (3) Neuron Structure Diversity Maximization (NEST-DM) loss to force the network to explore more features with discriminant ability as shown in Fig.~\ref{fig:Framework} (e).  


\subsection{NEuron STructure Modeling}



\noindent \textbf{NEuron STructure Representation.} For the rPPG measurement task, the convolution kernels are adaptively activated to extract physiological features against various noise distribution~\cite{liu2018frequency,wang2016cnnpack,tang2022defects}. However, the facial physiological signals are feeble and can be easily affected by various unseen noises, which make the filters activated abnormally.
Therefore, the NEuron STructure (NEST) representation is defined to explore how domain shift affects the model robustness in view of activated feature distribution. Specifically, we firstly average the feature map $m_{j}$ $\in$ $R^{W \times H \times C_{j}}$ over the spatial dimensions in the end of $j$-th layer to get the preliminary statistical presentation $\bar{o}_{j} \in R^{C_j}$ of the neuron structure, where $C_j$ is the channel number of the $j$-th layer. Following that, the preliminary presentation $\bar{o}_{j}$ is fed into the max-min normalized to get the NEST representation ${o_{j}} \in R^C_j$ of the $j$-th layer. The full NEST representation of $i$-th sample $x_i$ is defined as $O_i=\left\{o_{i,1},o_{i,2},...,o_{i,U}\right\}\in R^{(C_1 + C_2+ ...+ C_U)}$, where $U$ is the number of selected layer. 
\begin{figure}[!t]
\begin{center}
\includegraphics[scale=0.35]{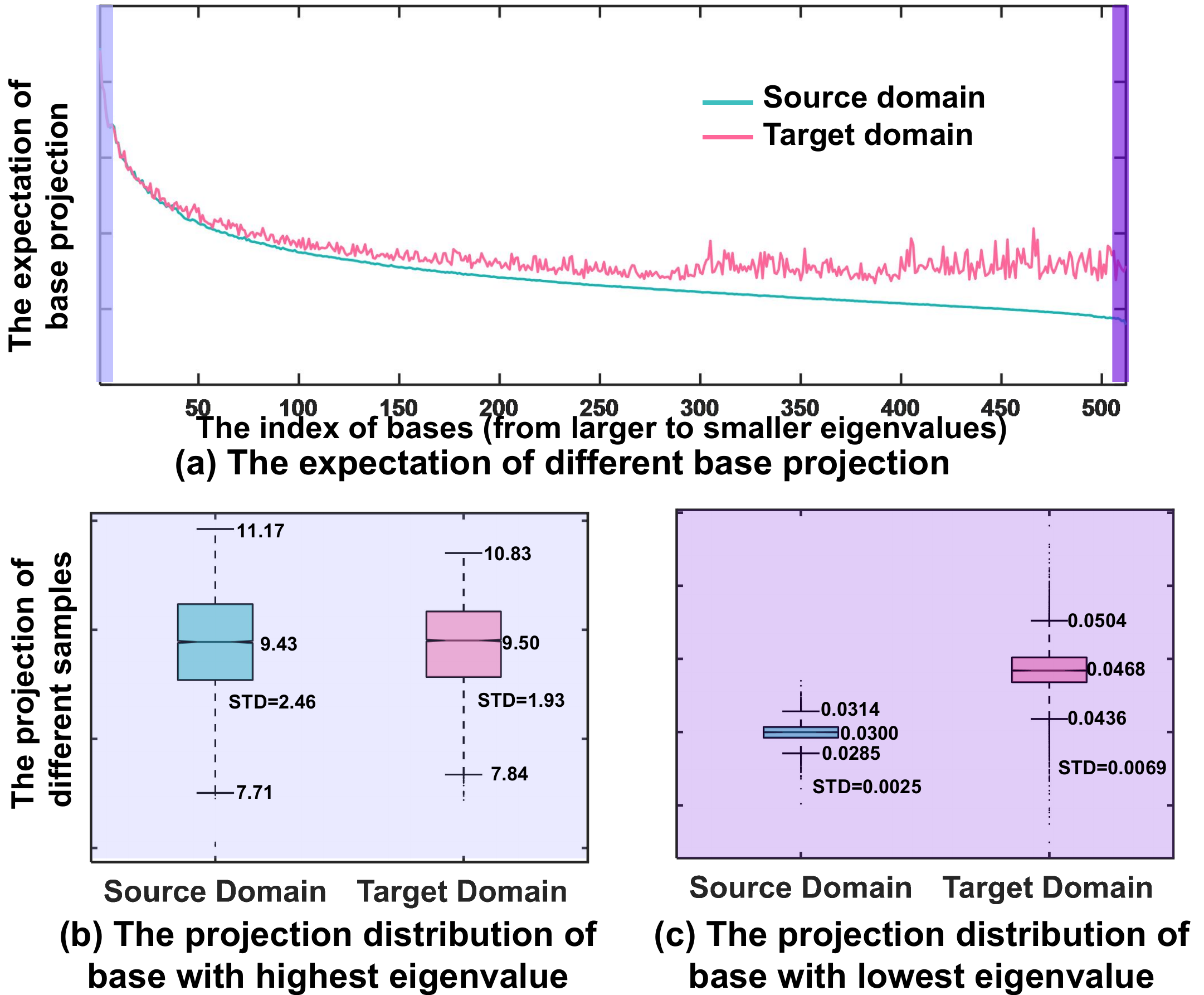}
\end{center}
\vspace{-5mm}
\caption{(a) The expectation of different base projection. (b) The distribution of base with highest eigenvalue. (c) The distribution of base with lowest eigenvalue.}
\label{fig:NEST-CM}
\vspace{-5mm}
\end{figure}

\noindent \textbf{NEuron STructure Coverage Maximization.} Existing works show that the OOD samples may activate abnormal neuron structures (feature combination) that seldom be trained with the source data so that the model's performance degrades drastically~\cite{NC1,NC2,NC2022PAMI}. To investigate this phenomenon in rPPG measurement task, we analyze the optimized degree of different feature combinations. 

Specifically, we decompose the NEST representation of the last layer into orthogonal bases with Singular Value Decomposition (SVD). Here, small eigenvalues indicate that training samples cover a small space in these corresponding bases. Thus, unseen target samples are more likely to lay out of the well-optimized space, causing a larger domain shift. To validate this, we plot the means of source and target samples on these bases in Fig.~\ref{fig:NEST-CM}(a). As shown, the mean differences are larger for bases of smaller eigenvalues. To make it clearer, We further visualize the data distribution on the first and last bases in Fig.~\ref{fig:NEST-CM} (b) and (c). As shown, the distribution difference in the last base is much more significant than that of the first one. All these suggest that for the NEST representation, components of small eigenvalues are less optimized during training, and domain shifts are more likely to appear there.

\vspace{-2mm}
\begin{figure}[!t]
\begin{center}
\includegraphics[width=1.0\linewidth]{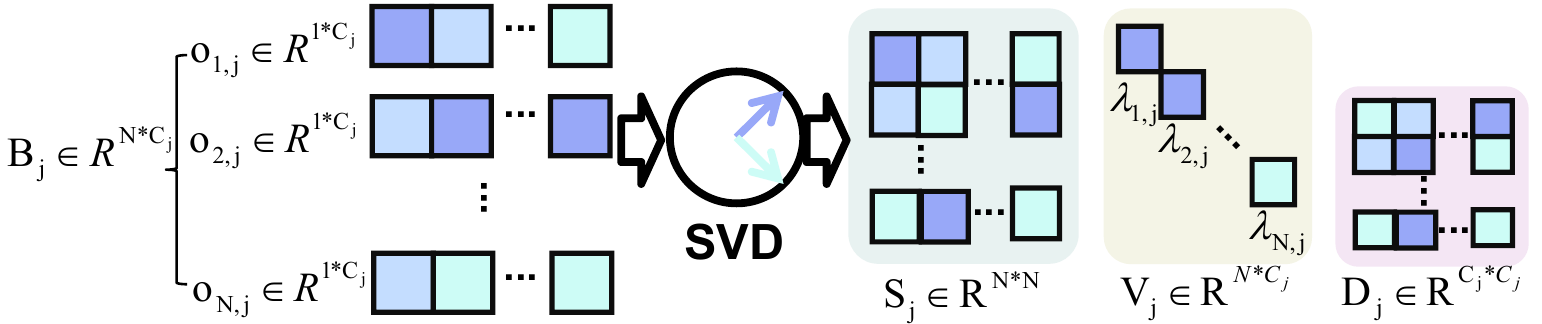}
\vspace{-7mm}
\caption{The detail of SVD for the $j$-th layer in CNN.}
\label{fig:SVD}
\vspace{-7mm}
\end{center}
\end{figure}

In sight of this, we propose the NEuron STructure Coverage Maximization (NEST-CM) regularization that improves the eigenvalues of the last few bases. This could narrow the under-optimized space as shown in Fig.~\ref{fig:Framework} (c).
For a training batch with $N$ samples, NEST representation set of $j$-th layer $B_{j}=\left\{o_{1,j}; o_{2,j}; ...; o_{N,j}\right\} \in R^{N*C_j}$ is decomposed to $S_j \cdot V_j \cdot D_j$ by SVD method as shown in Fig.~\ref{fig:SVD}, i.e., $B_j=S_j \cdot V_j \cdot D_j$, where $V_j \in R^{N*C_j}$ and $N$\textgreater $C_j$. The bases with small eigenvalue values are rare and under-optimized during the training stage. To reduce the occurrence probability of under-optimized NEST caused by domain shift, we stimulate under-optimized bases by maximizing the corresponding eigenvalues:  
\begin{equation}
\label{eq:NSM}
\mathcal{L}_{CM} = - \mathbb{E}_{j \in U} \frac{\sum_{i=1}^{C'} \lambda_{i,j}^{'}}{\sum_{i=1}^{C} \lambda_{i,j}}, {\lambda_{i,j}^{'}}<t, 
\end{equation} 
where $\lambda_{i,j}$ represent the $i$-th normalized eigenvalue in $j$-th layer' diagonal matrix $V_j$, and $\lambda_{i,j}^{'}$ is the eigenvalue less than the threshold $t$. Considering different channel numbers in each layer, we design an adaptive threshold $t$ to select the relatively under-optimized bases, i.e., $t=\frac{\rho}{C_j}$, where $\rho$ is a trade-off parameter. Besides, $C'_j$ is the number of eigenvalues of $j$-th layer under the threshold, and $U$ is the number of layers. 

\noindent \textbf{NEuron STructure Targeted Alignment.} Besides the NEST-CM introduced above, we further improve the generalization ability by aligning the NEST representation of different domains. Domain invariant learning has demonstrated its effectiveness in DG problems~\cite{wang2022DG,ganin2015AD,VREx,NC2022PAMI,GroupDRO,lv2022causality}. Some recent research~\cite{luo2021category,jiang2020implicit,deng2019cluster,wad2022equivariance} further shows that aligning domain features within the same labels could lead to better performance. However, labels in rPPG are continuous, and it is hard to find samples with identical labels in a training batch. 


To cope with this, we propose a novel regularization term called NEuron STructure Alignment (NEST-TA).
NEST-TA builds a synthesized prototype $O_i^p$ with the same physiological information as the given targeted sample $O_i$. Then, the feature within the same labels can be aligned across domains like~\cite{luo2021category,jiang2020implicit,deng2019cluster,wad2022equivariance}.
Specifically, we select the top $K$ samples closest to the targeted ground truth $y_{i}$, i.e., $[O'_{i,1}, ..., O'_{i,k}, ..., O'_{i,K}]$. The NEST center of the selected samples $\overline{O}'_{i}$ is calculated by averaging the NEST presentation of selected samples. Then we calculate the direction vectors between the NEST presentation center of the selected samples $\overline{O}'_{i}$ and the NEST presentation of each sample $O'_{i,k}$. Following that, the NEST center of the selected samples $\overline{O}'_{i}$ is calibrated to the synthesized prototype $O_i^p$, which is performed by using the Gaussian distribution function to fuse these direction vectors:
\vspace{-8mm}
\begin{center}
\begin{equation}
\begin{split}
O_i^p = \overline{O'}_{i} + \sum_{k=1}^K W_{i,k} * \frac{y_{i}-\overline{y}'_{i}}{y'_{i,k}-\overline{y}'_{i}} (O'_{i,k}&-\overline{O'}_{i}), \\ 
W_{i,k} = \frac{G(y_{i,k}-\overline{y}'_{i}, \sigma)}{\sum_{k=1}^K G(y_{i,k}-\overline{y}'_{i}, \sigma)},
\label{eq:proto}
\end{split}
\end{equation}
\end{center}
where $G(-, \sigma)$ is the Gaussian distribution with the mean of zero and the standard deviation of $\sigma$, i.e., $G(y_{i,k}-\overline{y}'_{i}, \sigma)= \frac{1}{\sqrt{2\pi}} e^{-\frac{(y_{i,k}-\overline{y}'_{i})^2}{2{\sigma}^2}}$, where $\sigma$ is a hyper-parameter. To align the feature distribution across different domains, NEST-TA uses the cosine similarity $Cos(-, -)$ to reduce the variant features between targeted samples $O_i$ and prototype sample $O_i^p$:
\vspace{-8mm}
\begin{center}
\begin{equation}
\begin{split}
\mathcal{L}_{TA} = \mathbb{E}_{i \in N} Cos(O_{i}, O_{i}^p).
\label{eq:NSDS}
 \end{split}
\end{equation}
\end{center}
\noindent \textbf{NEuron STructure Diversity Maximization.} Existing works elaborate that it is effective to explore more features with discriminant information to improve the generalization~\cite{lv2022causality,ren2022robust,chen2021scatterbrain,piratla2020efficient}. Motivated by this, NEuron STructure Diversity Maximization (NEST-DM) is introduced to increase the distinction between the NEST presentation of different samples as shown in Fig.~\ref{fig:Framework} (e). By doing this, the network is forced to explore more features with physiological information in a contrastive learning way, which can further improve model discrimination. NEST-DM $\mathcal{L}_{DM}$ can be represented as:

\vspace{-7mm}
\begin{center}
\begin{equation}
\begin{split}
\mathcal{L}_{DM} = - \mathbb{E}_{i \in N} log(\frac{e^{Cos(O_i,\widehat{O}_i)/\tau}}{\sum_{j=1(i\neq j)}^N e^{Cos(O_i,\widehat{O}_j)/\tau}}),
  \label{eq:DM}
 \end{split}
\end{equation}
\end{center}
where $\widehat{O}_i$ is the NEST representation of augmented sample from $O_i$, and $N$ and $\tau$ are the batch size and temperature parameter. The augmentation strategies for STMap will be elaborated in section~\ref{STMap}.
\vspace{-1.5mm}
\noindent \textbf{Model Training.} For the rPPG measurement task, we chose two popular losses in this field. Specifically, L1 loss $\mathcal{L}_{L1}$ and negative Pearson's correlation coefficient $\mathcal{L}_{p}$ are used to constrain the predicted heart rate value and the predicted BVP signal respectively~\cite{yu2019remote,CVD2020,lu2021dual,hu2021eta}. These two losses are also used for augmented data. With the violent confrontation at the beginning of training, $\mathcal{L}_{L1}$ make network tends to predict the mean of the heart rate value distribution instead of learning the physiological feature like $\mathcal{L}_{p}$. Besides, $\mathcal{L}_{NSM}$, $\mathcal{L}_{NEST-S}$, and $\mathcal{L}_{NEST-DM}$ also have meaningless effects on under-trained networks at the beginning. 

To solve the above-illustrated issues, we introduce the adaptation factor $\gamma = \frac{2}{1+exp(-10 r)}$ and $r = \frac{inters_{current}}{inters_{total}}$ to suppress meaningless constraints at the early training stage, which can accelerate model convergence and improve training stability\cite{ganin2015AD,jia2020single}. The final loss $\mathcal{L}_{overall}$ can be formulated as follows:
\vspace{-8mm}
\begin{center}
\begin{equation}
\begin{split}
\mathcal{L}_{overall} = k_1 \mathcal{L}_{p} + \gamma (k_2 \mathcal{L}_{L1} + k_3 \mathcal{L}_{CM} \\
+  k_4 \mathcal{L}_{TA} +  k_5 \mathcal{L}_{DM}),
  \label{eq:overall}
 \end{split}
\end{equation}
\end{center}
where $k_1$ - $k_5$ are five trade-off parameters. 



\section{Experiments}

\subsection{Datasets\label{Datasets}}
We elaborately select five datasets (including different motion, camera, and lighting conditions) to curate and benchmark a large-scale DG protocol:

\noindent \textbf{VIPL-HR}~\cite{niu2019robust} have nine scenarios, three RGB cameras, different illumination conditions, and different levels of movement. It is worth mentioning that the BVP signal and video of the dataset do not match in the time dimension, so $\mathcal{L}_{p}$ is not used for this dataset. We used the output of HR head as the evaluation result. Besides, we normalize STMap to 30 fps by cubic spline interpolation to solve the problem of the unstable frame rate of video~\cite{song2020heart,lu2021dual,yu2021physformer}.


\noindent \textbf{V4V}~\cite{revanur2021V4V} is designed to collect data with the drastic changes of physiological indicators by simulating ten tasks such as a funny joke, 911 emergency call, and odor experience. It is worth mentioning that There are only heart rate labels in the dataset and no BVP signal labels, so $\mathcal{L}_{p}$ is not used for this dataset. We used the output of HR head as the evaluation result.

\noindent \textbf{BUAA}~\cite{xi2020BUAA} is proposed to evaluate the performance of the algorithm against various illumination. We only use data with illumination greater than or equal to 10 lux because underexposed images require special algorithms that are not considered in this article. We used the output of BVP head as the evaluation for BUAA, PURE, and UBFC-rPPG datasets.

\noindent \textbf{PURE}~\cite{PURE2014} contains 60 RGB videos from 10 subjects with six different activities, specifically, sitting still, talking, and four rotating and moving head variations. The BVP signals are down-sampled from 60 to 30 fps with cubic spline interpolation to align with the videos.

\noindent \textbf{UBFC-rPPG}~\cite{UBFC2017} containing 42 face videos with sunlight and indoor illumination, the ground-truth BVP signals, and HR values were collected by CMS50E.

\begin{table*}[!ht]
\centering
\small
\setlength{\tabcolsep}{0.0001mm}
\begin{tabular}{cccccccccccccccc}
\toprule 
 & \multicolumn{3}{c}{\textbf{UBFC}}             & \multicolumn{3}{c}{\textbf{PURE}}             & \multicolumn{3}{c}{\textbf{BUAA}}             & \multicolumn{3}{c}{\textbf{VIPL}}              & \multicolumn{3}{c}{\textbf{V4V}}               \\
\cmidrule(lr){2-4} \cmidrule(lr){5-7} 
\cmidrule(lr){8-10} \cmidrule(lr){11-13}  \cmidrule(lr){14-16} 
\textbf{Method} & \textbf{MAE↓}  & \textbf{RMSE↓} & \textbf{\quad r↑ \quad}    & \textbf{MAE↓}  & \textbf{RMSE↓} & \textbf{\quad r↑ \quad}    & \textbf{MAE↓}  & \textbf{RMSE↓} & \textbf{\quad r↑ \quad}    & \textbf{MAE↓}  & \textbf{RMSE↓}  & \textbf{\quad r↑ \quad}    & \textbf{MAE↓}  & \textbf{RMSE↓}  & \textbf{\quad r↑ \quad}    \\
\midrule 
\textbf{GREEN~\cite{verkruysse2008remote_GREEN}}                                              & 8.02          & 9.18          & 0.36          & 10.32         & 14.27         & 0.52          & 5.82          & 7.99          & 0.56          & 12.18         & 18.23          & 0.25          & 15.64         & 21.43          & 0.06          \\
\textbf{CHROM~\cite{de2013robust_CHROM}}                                              & 7.23          & 8.92          & 0.51          & 9.79          & 12.76         & 0.37          & 6.09          & 8.29          & 0.51          & 11.44         & 16.97          & 0.28          & 14.92         & 19.22          & 0.08          \\
\textbf{POS~\cite{wang2017algorithmic_POS}}                                                & 7.35          & 8.04          & 0.49          & 9.82          & 13.44         & 0.34          & 5.04          & 7.12          & 0.63          & 14.59         & 21.26          & 0.19          & 17.65         & 23.22          & 0.04          \\
\midrule
\textbf{DeepPhys~\cite{Chen_2018_ECCV}}                                           & 7.82          & 8.42          & 0.54          & 9.34          & 12.56         & 0.55          & 4.78          & 6.74          & 0.69          & 12.56         & 19.13          & 0.14          & 14.52         & 19.11          & 0.14          \\
\textbf{TS-CAN~\cite{liu2020multi}}                                             & 7.63          & 8.25          & 0.55          & 9.12          & 12.38         & 0.57          & 4.84          & 6.89          & 0.68          & 12.34         & 18.94          & 0.16          & 14.77         & 19.96          & 0.12          \\
\textbf{Rhythmnet$^*$~\cite{niu2019robust}}                                           & 5.79          & 7.91          & 0.78          & 7.39          & 10.49         & 0.77          & 3.38          & 5.17          & 0.84          & 8.97          & 12.16          & 0.49          & 10.16         & 14.57          & 0.34          \\
\textbf{Dual-GAN$^*$~\cite{lu2021dual}}                                           & 5.55          & 7.62          & 0.79          & 7.24          & 10.27         & 0.78          & 3.41          & 5.23          & 0.84          & 8.88          & 11.69          & 0.50          & 10.04         & 14.44          & 0.35          \\
\textbf{BVPNet$^*$~\cite{bvpnet}}                                           & 5.43          & 7.71          & 0.80          & 7.23           & 10.25         & 0.78          & 3.69          & 5.48          & 0.81          & 8.45          & 11.64          & 0.51          & 10.01         & 14.35          & 0.36          \\
\midrule
\textbf{AD$^{*+}$~\cite{ganin2015AD}}                                                 & 5.92          & 8.08          & 0.76          & 7.42          & 10.61         & 0.73          & 3.49          & 5.49          & 0.82          & 8.41          & 11.71          & 0.53          & 10.47         & 14.64          & 0.32          \\
\textbf{GroupDRO$^{*+}$~\cite{GroupDRO}}                                           & 5.73          & 7.97          & 0.78         & 7.69          & 10.83         & 0.78          & 3.41          & 5.21          & 0.83             & 8.35          & 11.67          & 0.54          & 9.94          & 14.29          & 0.36          \\

\textbf{Coral$^{*+}$~\cite{coral}}                                           & 5.89          & 8.04          & 0.76          & 7.59          & 10.87         & 0.72          & 3.64          & 5.74          & 0.80          & 8.68          & 11.91          & 0.53          & 10.32         & 14.42          & 0.32      \\

\textbf{VREx$^{*+}$~\cite{VREx}}                                               & 5.59          & 7.68          & 0.81      & 7.24          & 10.14         & 0.78         & 3.27          & 5.01          & 0.86             & 8.37          & 11.62          & 0.54          & 9.82          & 14.16          & 0.37          \\
\textbf{NCDG$^{*+}$~\cite{NC2022PAMI}}                                               & 5.31          & 7.56          & 0.82          & 7.32          & 10.35         & 0.77          & 3.12          & 5.16          & 0.85          & 8.47          & 11.81          & 0.52          & 10.14         & 14.46          & 0.34          \\
\midrule
\textbf{Baseline$^*$~\cite{niu2019robust}}                                           & 5.79          & 7.91          & 0.78          & 7.39          & 10.49         & 0.77          & 3.38          & 5.17          & 0.84          & 8.97          & 12.16          & 0.49          & 10.16         & 14.57          & 0.34          \\
\textbf{NEST$^{*+}$ w/o $\mathcal{L}_{CM}$}  & 4.92          & 7.19          & 0.83          & 6.91          & 9.94          & 0.78          & \textbf{2.79} & \textbf{4.62} & \textbf{0.89} & 8.21          & 11.56          & 0.55          & 9.58          & 13.58          & 0.38          \\
\textbf{NEST$^{*+}$ w/o $\mathcal{L}_{TA}$} & 4.97          & 7.28          & 0.83          & 6.81          & 9.77          & 0.81          & 3.03          & 4.94          & 0.86          & 8.19          & 11.47          & 0.55          & 9.52          & 13.52          & 0.38          \\
\textbf{NEST$^{*+}$ w/o $\mathcal{L}_{DM}$}  & 4.89          & 7.11          & 0.84          & 6.87          & 9.81          & 0.79          & 2.93          & 4.77          & 0.88          & 8.14          & 11.43          & 0.56          & 9.49          & 13.49          & 0.40           \\
\textbf{NEST$^{*+}$}                                            & \textbf{4.67} & \textbf{6.79} & \textbf{0.86} & \textbf{6.71} & \textbf{9.59} & \textbf{0.81} & 2.88          & 4.69          & 0.89          & \textbf{7.86} & \textbf{11.15} & \textbf{0.58} & \textbf{9.27} & \textbf{13.79} & \textbf{0.41} \\
\bottomrule 
\end{tabular}
\vspace{-2mm}
\caption{HR estimation results on MSDG protocol. It is worth noting that we have removed the GRU mechanism in Rhythmnet because it is an unfair comparison with using GRU to aggregate the temporal information of multiple clips instead of only one clip; $^{*}$ means that these methods use the STMap as the input of CNN and use the STMap augmentation strategy as mentioned in section~\ref{STMap}; $^{+}$ means that these methods are based on baseline (Rhythmnet without GRU).}
\label{HR_DG}
\vspace{-5mm}
\end{table*}

\vspace{-1.5mm}
\subsection{Implementation Details}
\label{STMap}
\vspace{-1.5mm}
\noindent 
\textbf{Training Details.} Our proposed method is performed under the Pytorch framework. For the generation of STMap from Video, we use the FAN~\footnote{\url{https://github.com/1adrianb/face-alignment}} to detect the 2D landmarks of faces~\cite{bulat2017far}, and other steps completely follow~\cite{niu2019robust}. STMap in each dataset is sampled with a time window of 256 with step 10. 

Following~\cite{sun2022ECCV,wang2022selfAAAI}, we also design spatial and temporal augmentation for STMap, i.e., shuffle each row of STMap and slide the time window. In addition, color jitter and blur augmentation are used for STMap~\cite{chen2020simple}. Then the STMap $x \in R^{25\times256\times3}$ is resized to $x' \in R^{64\times256\times3}$ as the input of the network. We use the ResNet-18 with removing the max pooling layer as the feature extractor, and the final full connection layer is used as the HR estimation head. At the same time, we designed a BVP signal regression head containing four blocks, each block involves a Conv., transpose, and two Conv. layers with batch normalization and Relu function. Hyper-parameter $\sigma$, $\rho$ and $\tau$ are set to 5, 0.1, and 0.2 according to the experimental result, and the trade-off parameter $k_1 - k_5$ are set to 1, 0.1, 0.001, 0.1 and 0.01 based on the scale of losses. Adam optimizer with a learning rate of 0.001 is used for training. The batch size and iterations are set to 1024 and 20000.

\noindent \textbf{Performance Metrics.}  Following existing methods~\cite{Chen_2018_ECCV,song2020heart,yu2021physformer,CVD2020,lu2021dual}, standard deviation (SD), mean absolute error (MAE), root mean square error (RMSE), and Pearson’s correlation coefficient (r) are used to evaluate the HR estimation. MAE, RMSE, and r are used to evaluate the HRV measurement, including low frequency (LF), high frequency (HF), and LF/HF.

\subsection{Multi-Source Domain Generalization}
\vspace{-1mm}
\subsubsection{HR Estimation}
\vspace{-1.5mm}
For remote HR estimation, all five datasets (i.e., UBFC~\cite{UBFC2017}, PURE~\cite{PURE2014}, BUAA~\cite{xi2020BUAA}, VIPL~\cite{niu2019robust}, and V4V~\cite{revanur2021V4V}) are used to evaluate the model generalization. Specifically, four datasets are used for training, and then the other dataset is used to test. Firstly, we performance the iPhys~\footnote{\url{https://github.com/danmcduff/iphys-toolbox}} to evaluate three traditional method (i.e., GREEN~\cite{verkruysse2008remote_GREEN}, CHROM~\cite{de2013robust_CHROM}, and POS~\cite{wang2017algorithmic_POS}) as shown in Tab.~\ref{HR_DG}. Then, we reproduce five popular DL-based methods (i.e., DeepPhys~\cite{Chen_2018_ECCV}, TS-CAN~\cite{liu2020multi}, Rhythmnet~\cite{niu2019robust}, and BVPNet~\cite{bvpnet}, and Dual-GAN~\cite{lu2021dual}) for comparison on the same setting. Subsequently, we directly applied the four DG methods (i.e., AD~\cite{ganin2015AD}, GroupDRO~\cite{GroupDRO}, Coral~\cite{coral}, VREx~\cite{VREx}, and NCDG~\cite{NC2022PAMI}) to Rhythmnet with using DeepDG~\footnote{\url{https://github.com/jindongwang/transferlearn -ing/tree/master/code/DeepDG}\label{DeepDG}}. 

\noindent \textbf{The performance of traditional methods.} As shown in Tab.~\ref{HR_DG}, traditional algorithms show relatively acceptable results on some simple scenes, such as BUAA and UBFC, but very poor results on other datasets. It is difficult for these traditional algorithms to extract the physiological information on the V4V dataset, which shows that the traditional methods cannot deal with the noise caused by the complex environment and non-rigid motions. 

\begin{table*}[]
\setlength{\tabcolsep}{0.8mm}
\centering
\small
\begin{tabular}{clcccccccccccc}
\toprule
 &   & \multicolumn{3}{c}{\textbf{LF-(u.n)}}   & \multicolumn{3}{c}{\textbf{HF-(u.n)}}   & \multicolumn{3}{c}{\textbf{LF/HF}} & \multicolumn{3}{c}{\textbf{HR-(bpm)}}     \\
\cmidrule(lr){3-5} \cmidrule(lr){6-8} 
\cmidrule(lr){9-11} \cmidrule(lr){12-14}        
{\textbf{Target}}&{\textbf{Method}}  & \textbf{MAE↓}    & \textbf{RMSE↓}   & \textbf{\quad r↑ \quad}      & \textbf{MAE↓}    & \textbf{RMSE↓}   & \textbf{\quad r↑ \quad}     & \textbf{MAE↓}    & \textbf{RMSE↓}   & \textbf{\quad r↑ \quad}     & \textbf{MAE↓}    & \textbf{RMSE↓}   & \textbf{\quad r↑ \quad}     \\
\midrule 
\textbf{UBFC} & 
\textbf{GREEN~\cite{verkruysse2008remote_GREEN}}    & 0.2355 & 0.2841 & 0.0924 & 0.2355 & 0.2841 & 0.0924 & 0.6695  & 0.9512 & 0.0467 & 8.0184  & 9.1776  & 0.3634 \\
& \textbf{CHROM~\cite{de2013robust_CHROM}}    & 0.2221 & 0.2817 & 0.0698 & 0.2221 & 0.2817 & 0.0698 & 0.6708  & 1.0542 & 0.1054 & 7.2291  & 8.9224  & 0.5123 \\
& \textbf{POS~\cite{wang2017algorithmic_POS}}      & 0.2364 & 0.2861 & 0.1359 & 0.2364 & 0.2861 & 0.1359 & 0.6515  & 0.9535 & 0.1345 & 7.3539  & 8.0402  & 0.4923 \\
& \textbf{Baseline} & 0.0621 & 0.0813 & 0.1873 & 0.0621 & 0.0813 & 0.1873 & 0.1985  & 0.2667 & 0.3043 & 5.1542  & 7.4672  & 0.8165 \\
& \textbf{NEST}  & \textbf{0.0597} & \textbf{0.0782} & \textbf{0.2017} & \textbf{0.0597} & \textbf{0.0782} & \textbf{0.2017} & \textbf{0.2138}  & \textbf{0.2824} & \textbf{0.3179} & \textbf{4.7471}  & \textbf{6.8876}  & \textbf{0.8546} \\
 \midrule 
\textbf{PURE} &\textbf{GREEN~\cite{verkruysse2008remote_GREEN}}   & 0.2539 & 0.3002 & 0.0326 & 0.2539 & 0.3002 & 0.0326 & 0.6525  & 0.8932 & 0.0417 & 10.3247 & 14.2693 & 0.4952 \\
& \textbf{CHROM~\cite{de2013robust_CHROM}}   & 0.2096 & 0.2751 & 0.1059 & 0.2096 & 0.2751 & 0.0759 & 0.5404  & 0.8266 & 0.1173 & 9.7914  & 12.7568 & 0.3732 \\
& \textbf{POS~\cite{wang2017algorithmic_POS}}      & 0.1959 & 0.2571 & 0.1684 & 0.1959 & 0.2571 & 0.1684 & 0.5373  & 0.846  & 0.1433 & 9.8273  & 13.4414 & 0.3432 \\
& \textbf{Baseline} & 0.0671 & 0.0923 & 0.6046 & 0.0671 & 0.0923 & 0.6046 & 0.2864  & 0.4184 & 0.5526 & 8.2542  & 11.1765 & 0.6832 \\
& \textbf{NEST}  & \textbf{0.0635} & \textbf{0.0874} & \textbf{0.6422} & \textbf{0.0635} & \textbf{0.0874} & \textbf{0.6422} & \textbf{0.2255}  & \textbf{0.3505} & \textbf{0.5734} & \textbf{7.6889}  & \textbf{10.4783} & \textbf{0.7255} \\
  \midrule                     
\textbf{BUAA} & \textbf{GREEN~\cite{verkruysse2008remote_GREEN}}    & 0.3472 & 0.3951 & 0.0871 & 0.3472 & 0.3951 & 0.0871 & 0.6453  & 0.8632 & 0.0921 & 5.8231  & 7.9882  & 0.5624 \\
                      & \textbf{CHROM~\cite{de2013robust_CHROM}}    & 0.3786 & 0.3237 & 0.0682 & 0.3786 & 0.3237 & 0.0682 & 0.6813  & 0.8836 & 0.0715 & 6.0934  & 8.2938  & 0.5165 \\
                      & \textbf{POS~\cite{wang2017algorithmic_POS}}      & 0.3198 & 0.3762 & 0.0962 & 0.3198 & 0.3762 & 0.0962 & 0.6275  & 0.8424 & 0.1127 & 5.0407  & 7.1198  & 0.6374 \\
                      & \textbf{Baseline} & 0.1451 & 0.1681 & 0.2891 & 0.1451 & 0.1681 & 0.2891 & 0.5564  & 0.6904 & 0.2914 & 3.7852  & 6.3237  & 0.7492 \\
                      & \textbf{NEST}  & \textbf{0.1436} & \textbf{0.1665} & \textbf{0.2955} & \textbf{0.1436} & \textbf{0.1665} & \textbf{0.2955} & \textbf{0.5514}  & \textbf{0.6884} & \textbf{0.3004} & \textbf{3.3723}  & \textbf{5.8806}  & \textbf{0.7647} \\
\bottomrule 
\end{tabular}
\vspace{-2mm}
\caption{HRV and HR estimation results on the MSDG protocol.}
\label{HRV_DG}
\vspace{-2mm}
\end{table*}

\noindent \textbf{The performance of DL-based methods.} For DL-based methods, DeepPhys and TS-CAN even perform worse than traditional methods. Rhythmnet, BVPNet, and Dual-GAN perform much better than both other DL-based methods and traditional methods, which suggests that STMap is the more robust input of CNN than the aligned video. Compared with Rhythmnet, BVPNet and Dual-GAN have not significant improvement on the MSDG protocol. Because BVPNet and Dual-GAN consider the performance of the single domain instead of focusing on solving domain shift. Therefore, Rhythmnet is also selected as a baseline to further analyze the generalization on MSDG protocol.

\noindent \textbf{The performance of DG methods.} To further improve the generalization of the model, we apply the existing popular DG methods to baseline (Rhythmnet) as shown in Tab.~\ref{HR_DG}. There is a slight model degradation of AD~\cite{ganin2015AD}, Coral~\cite{coral}, and GroupDRO~\cite{GroupDRO} on some target datasets. The reason behind this is that these algorithms employ inaccurate domain labeling, specifically, a dataset often with different acquisition devices, lighting, and scene cannot be considered as the same domain. On the contrary, VREx~\cite{VREx} and NCDG~\cite{NC2022PAMI} can slightly improve the performance of Rhythmnet without utilizing the domain label. Inaccurate domain division will make it difficult for the DG algorithm to effectively align the feature distribution across different domains, which is termed as \textit{agnostic domain generalization}. 

Without the domain labels, NEST experimentally manifested significant improvement on MSDG protocol. For the target dataset BUAA, the results without NEST-CM $\mathcal{L}_{CM}$ show the best performance. Because NEST-CM may cause the model degradation with activating meaningless features at the instance level, especially for the BUAA with the slight domain shift. 

\subsubsection{HRV Estimation}
We also conduct experiments for HRV (i.e., LF, HF, and LF/HF) and HR measurement on the PURE, UBFC, and BUAA. The reason for choosing only these three datasets is that VIPL and V4V do not have reliable ground-truth BVP signals, as mentioned in subsection~\ref{Datasets}. We also perform three tradition method (GREEN~\cite{verkruysse2008remote_GREEN}, CHMROM~\cite{de2013robust_CHROM}, POS~\cite{wang2017algorithmic_POS}) by iPhys for comparison with our method as shown in Tab.~\ref{HRV_DG}. With a more robust NEST presentation distribution, our method also can get better performance than baseline and other traditional methods in all metrics. 

\begin{table}[]
\small
\setlength{\tabcolsep}{0.1mm}
\centering
\begin{tabular}{ccccccc}
\toprule 
   & \multicolumn{3}{c}{\textbf{PURE}}             & \multicolumn{3}{c}{\textbf{BUAA}}             \\
\cmidrule(lr){2-4} \cmidrule(lr){5-7} 
\textbf{Method}   & \textbf{MAE↓}  & \textbf{RMSE↓} & \textbf{\quad r↑ \quad}    & \textbf{MAE↓}  & \textbf{RMSE↓} & \textbf{\quad r↑ \quad}    \\
 \midrule
\textbf{GREEN~\cite{verkruysse2008remote_GREEN}}    & 10.32         & 14.27         & 0.52          & 5.82          & 7.99          & 0.56          \\
\textbf{CHROM~\cite{de2013robust_CHROM}}    & 9.79          & 12.76         & 0.37          & 6.09          & 8.29          & 0.51          \\
\textbf{POS~\cite{wang2017algorithmic_POS}}      & 9.82          & 13.44         & 0.34          & 5.04          & 7.12          & 0.63          \\
  \midrule 
\textbf{Baseline} & 7.54          & 10.13         & 0.73          & 2.97          & 3.28          & 0.76          \\
\textbf{NEST}  & \textbf{6.07} & \textbf{9.06} & \textbf{0.76} & \textbf{2.56} & \textbf{2.73} & \textbf{0.78} \\
\bottomrule
\end{tabular}
\vspace{-2mm}
\caption{HR estimation results on SSDG protocol by training on the UBFC-rPPG.}
\label{HR_SSDG}
\vspace{-6mm}
\end{table}

\subsection{Single-Source Domain Generalization}
For a worst-case scenario, the model is trained on only one dataset to generalize to the unseen domains, namely single-source domain generalization (SSDG). Based on this, we select UBFC-rPPG as the source domain and then select PURE and BUAA as the target domain to further evaluate the generalization of NEST as shown in Tab.~\ref{HR_SSDG}. As we can see, NEST achieves desired performance by training on only one dataset. The reason is that: (1) NEST-CM can avoid the appearance of abnormal NEST; (2) NEST-TA can find the shared and robust feature of the samples within the same label; (3) NEST-DM can distinguish the boundaries between different samples to improve the generalization.

\subsection{Intra-Dataset Testing on VIPL}  
Our proposed method can also improve the effectiveness of the model on intra-dataset testing by reducing the harmful effects of abnormal disturbances. Following ~\cite{niu2019robust,CVD2020,lu2021dual,yu2021physformer}, we test NEST by using a 5-fold cross-validation protocol on VIPL-HR. Our method compared with ten baseline methods, including the traditional methods (SAMC~\cite{tulyakov2016self}, POS~\cite{wang2017algorithmic_POS} and CHROM~\cite{de2013robust_CHROM}) and the DL-based methods (I3D~\cite{CarreiraQuo}, DeepPhy~\cite{Chen_2018_ECCV}, RhythmNet~\cite{niu2019robust}, CVD~\cite{CVD2020}, BVPNet~\cite{bvpnet}, Physformer~\cite{yu2021physformer}, and Dual-GAN~\cite{lu2021dual}). The results of these methods are directly from~\cite{yu2021physformer,lu2021dual} as shown in Tab.~\ref{VIPL}.

\begin{table}[!t] 
\small
\centering
\begin{tabular}{lcccccc} 
\toprule  
\textbf{Method} & \textbf{SD↓}& \textbf{MAE↓}& \textbf{RMSE↓} & \textbf{\quad r↑ \quad} \\
  \midrule 
  {\textbf{SAMC}~\cite{tulyakov2016self}}  & 18.0 & 15.9 & 21.0  & 0.11 \\
  \textbf{POS}~\cite{wang2017algorithmic_POS}  & 15.3 & 11.5 & 17.2   & 0.30 \\
  \textbf{CHROM}~\cite{de2013robust_CHROM}& 15.1 & 11.4 & 16.9  & 0.28 \\
  \textbf{I3D}~\cite{CarreiraQuo}  & 15.9 & 12.0 & 15.9  & 0.07 \\
  \textbf{DeepPhy}~\cite{Chen_2018_ECCV} & 13.6 & 11.0 & 13.8  & 0.11 \\
  \textbf{BVPNet}~\cite{das2021bvpnet} & 7.75 & 5.34 & 7.85   & 0.70 \\
  \textbf{RhythmNet}~\cite{niu2019robust} & 8.11 & 5.30 & 8.14  & 0.76 \\
  \textbf{CVD}~\cite{CVD2020}  & 7.92 & 5.02 & 7.97  & 0.79 \\
  \textbf{Physformer}~\cite{yu2021physformer}  & 7.74 & 4.97& 7.79  & 0.78\\
  \textbf{Dual-GAN}~\cite{lu2021dual}  & 7.63 & 4.93& 7.68  & 0.81\\
  \midrule 
  \textbf{Baseline}  & 8.04 & 5.21 & 8.07   & 0.77 \\
  \textbf{NEST}  & \textbf{7.49}  & \textbf{4.76}  & \textbf{7.51}  &\textbf{0.84} \\
\bottomrule 
\end{tabular} 
\vspace{-1.4mm}
\caption{HR estimation results by our method and several state-of-the-art methods on the VIPL-HR database.}
\label{VIPL}
\end{table}

As shown in Tab.~\ref{VIPL}, the proposed methods outperform all the methods under all metrics. Compared with the baseline, it also gets a noticeable promotion, i.e., 0.4 in MAE and 0.53 in RMSE. It is worth noting that the VIPL dataset has multiple complex scenes and recording devices and cannot be considered as a single domain, termed the agnostic domain problem. This proves that NEST can effectively alleviate the domain shift without domain labels.

\subsection{Further Analysis}
We conduct further experiments on MSDG protocol (testing on UBFC with training on the other four datasets).

\begin{figure}[!t]
\centering
\begin{center}
\vspace{-4mm}
\includegraphics[scale=0.44]{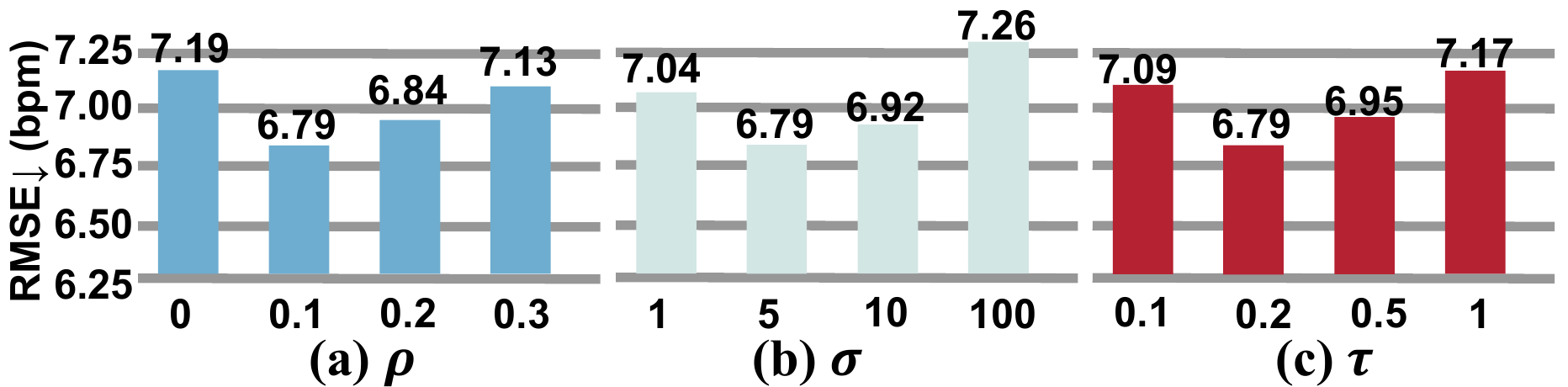}
\vspace{-5mm}
\caption{Impacts of the hyperparameter (a) $\rho$, (b) $\sigma$, and (c) $\tau$ of our proposed method.}
\label{fig:FA_1}
\vspace{-8mm}
\end{center}
\end{figure}

\noindent \textbf{Impact of $\rho$ in NEST-CM.} The hyperparameter $\rho$ determines how many bases need to be strengthened by NEST-CM $\mathcal{L}_{CM}$. As illustrated in Fig.~\ref{fig:FA_1} (a), NEST could achieve smaller RMSE when $\rho=0.1$, indicating that the bases with small eigenvalue require more attention and fine-tuning.

\noindent \textbf{Impact of $\sigma$ in NEST-TA.} The hyperparameter $\sigma$ tradeoffs the contribution of each selected sample to form the synthesized prototype. As illustrated in Fig.~\ref{fig:FA_1}, NEST could achieve the best performance with $\sigma=5$, which shows that the appropriate fusion of the nearby sample is conducive to improving model generalization. The specific reason behind this is that: (1) the synthesized prototype based on multi-domain samples can be used to enhance the domain invariant feature by NEST-TA; (2) NEST-TA can effectively aggregate the label relationship into NEST presentation distribution, which is performed by improving the similarity of NEST representation between the adjacent label samples.

\begin{figure}[!t]
\centering
\begin{center}
\includegraphics[scale=0.27]{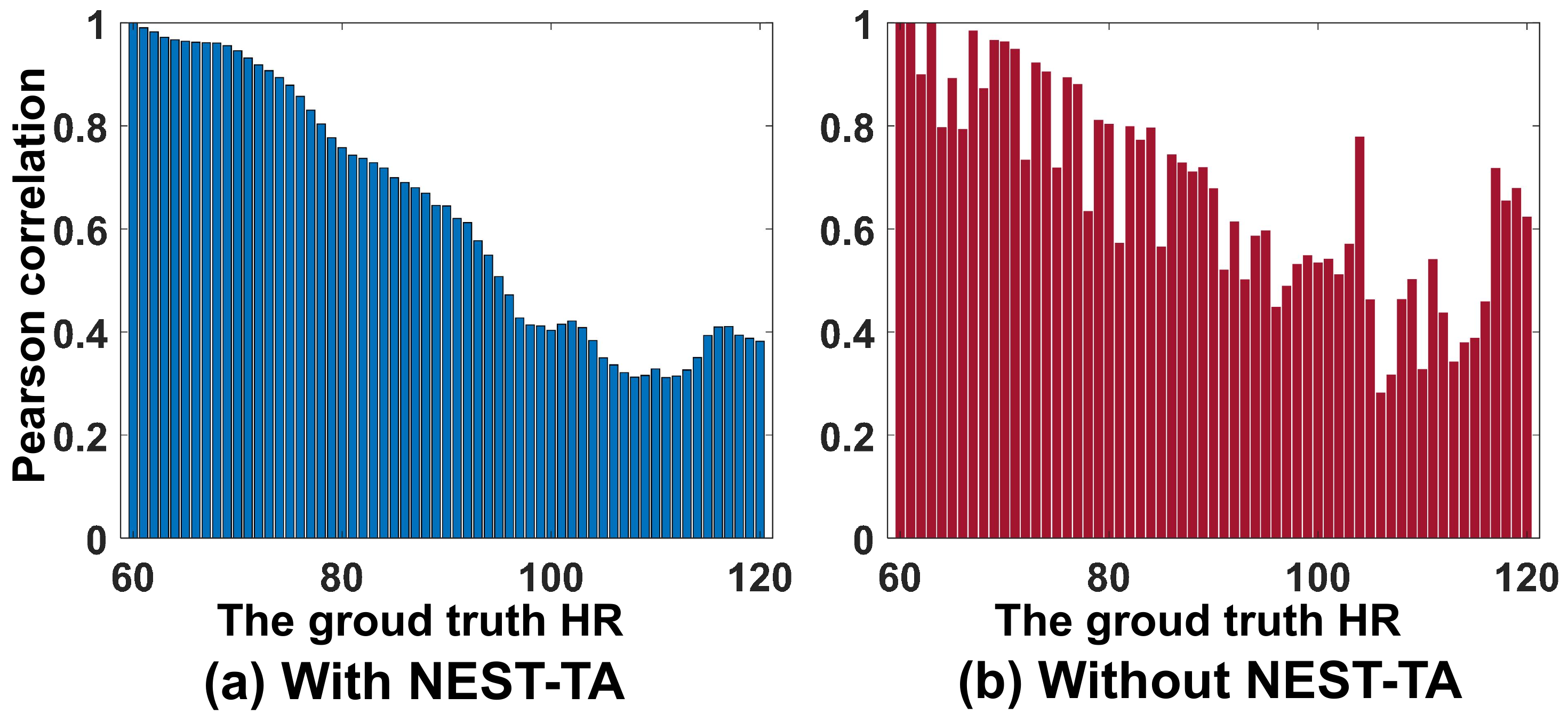}
\vspace{-4mm}
\caption{The pearson's correlation coefficient of the NEST representation of the samples with different labels for the anchor HR 60.}
\label{fig:NSDS}
\vspace{-10mm}
\end{center}
\end{figure}

To further clarify the second reason, we calculate the Pearson's correlation coefficient of the NEST representation of the samples with different labels for the anchor HR 60, as shown in Fig.~\ref{fig:NSDS}. It is obvious that the distribution changes with NEST-TA loss are more smooth across the continuous HR labels, which allows OOD samples with slight domain shifts to predict nearby heart rates around the ground truth HR value instead of other abnormal HR values.


\noindent \textbf{Impact of $\tau$ in NEST-DM.} The hyperparameter $\tau$ determines the pushing strength of the different samples. With the smallest $\tau$, NEST-DM $\mathcal{L}_{DM}$ has the overemphasized constraints to distinguish the NEST representation between different samples, which causes a harmful impact on the discriminant ability of the model. Overall, NEST-DM $\mathcal{L}_{DM}$ is an effective way to improve the generalization of model.



\vspace{-1mm}
\section{Conclusion}
\vspace{-1mm}

This paper presents the NEural STructure (NEST) modeling, a novel framework that addresses the domain generalization problem in rPPG measurement for the first time. Our framework contains three regularization terms defined in the view of neural structure, which improves the generalization capacity in the different aspects. All three terms do not rely on domain labels, which are especially suitable in rPPG, as defining domains is difficult in this task. To compare the generalization of the existing algorithm and our method, we establish the first DG protocol for this task. We experimentally manifested the substantial improvement of our method on both MSDG and SSDG protocols.

{\small
\bibliographystyle{ieee_fullname}
\bibliography{egbib}
}

\end{document}